\documentclass[letterpaper, 10 pt, conference]{IEEEtran}  %
\usepackage[letterpaper,left=.75in,right=.75in,top=.75in,bottom=.75in]{geometry}
\IEEEoverridecommandlockouts    %

\usepackage[pdftex]{graphicx}
\graphicspath{{figures/}}
\usepackage{hyperref}
\hypersetup{
    colorlinks=true,
    linkcolor=black,
    citecolor=black,
    filecolor=black,
    urlcolor=black,
}
\usepackage[T1]{fontenc}
\usepackage[utf8]{inputenc}
\usepackage{csquotes}
\usepackage[english]{babel}
\usepackage[export]{adjustbox}
\usepackage{caption}
\usepackage{subcaption}
\usepackage[colorinlistoftodos]{todonotes} %
\usepackage{lipsum}
\renewcommand\lipsum[1][2]{lorem ipsum} %

\usepackage{placeins}%

\usepackage{amsmath} %
\usepackage{fdsymbol} %
\usepackage{bm} %
\usepackage{siunitx} %
\usepackage{cleveref}
\crefname{equation}{}{}
\Crefname{equation}{}{}
\crefrangelabelformat{equation}{(#3#1#4)--(#5#2#6)}
\crefmultiformat{equation}{(#2#1#3)}{ and~(#2#1#3)}{, (#2#1#3)}{ and~(#2#1#3)}
\crefname{figure}{Fig.}{Fig.}
\Crefname{figure}{Fig.}{Fig.}
\crefmultiformat{figure}{Fig.~#2#1#3}{ and~#2#1#3}{, #2#1#3}{ and~#2#1#3}
\crefrangelabelformat{figure}{#3#1#4--#5#2#6}
\crefformat{footnote}{#2\footnotemark[#1]#3}

\usepackage[style=ieee,
            doi=false,
            url=false,
            mincitenames=1,
            maxcitenames=1,
            minbibnames=6,
            maxbibnames=6,
            backend=biber]{biblatex}  %
\newcommand{\kmh}{\tfrac{\kilo\meter}{\hour}}
\newcommand{\mps}{\tfrac{\meter}{\second}}
\newcommand{\mpss}{\tfrac{\meter}{\second^2}}

\newcommand{\dt}{{\scriptstyle\Delta}t}

\title{\vspace{.25in}
Tackling Occlusions \& Limited Sensor Range with Set-based Safety Verification
}

\author{\IEEEauthorblockN{Piotr F. Orzechowski\IEEEauthorrefmark{1},
Annika Meyer\IEEEauthorrefmark{1},
Martin Lauer\IEEEauthorrefmark{2}}
\IEEEauthorblockA{\IEEEauthorrefmark{1}Intelligent Systems and Production Engineering,
FZI Research Center for Information Technology,
Karlsruhe, Germany\\
Email: \{orzechowski,ameyer\}@fzi.de}
\IEEEauthorblockA{\IEEEauthorrefmark{2}Institute of Measurement and Control Systems,
Karlsruhe Institute of Technology (KIT),
Karlsruhe, Germany\\
Email: martin.lauer@kit.edu}}

\begin{document}

\maketitle

\IEEEpubid{\begin{minipage}{\textwidth}~\\[12pt] \centering%
   10.1109/ITSC.2018.8569332~ %
  \copyright~2018 IEEE. Personal use of this material is permitted. Permission from IEEE must be obtained for all other uses, including reprinting/republishing this material for advertising or promotional purposes, collecting new collected works for resale or redistribution to servers or lists, or reuse of any copyrighted component of this work in other works.
\end{minipage}}
\IEEEpubidadjcol

\pagestyle{empty}

\begin{abstract}

Provable safety is one of the most critical challenges in automated driving.
The behavior of numerous traffic participants in a scene cannot be predicted reliably
due to complex interdependencies and the indiscriminate behavior of humans.
Additionally, we face high uncertainties and only incomplete environment knowledge.
Recent approaches minimize risk with probabilistic and machine learning methods -- even under occlusions.
These generate comfortable behavior with good traffic flow, but cannot guarantee safety of their maneuvers.
Therefore, we contribute a safety verification method for trajectories under occlusions.
The field-of-view of the ego vehicle and a map are used to identify critical sensing field edges,
each representing a potentially hidden obstacle. %
The state of occluded obstacles is unknown, but can be over-approximated by intervals over all possible states.
Then set-based methods are extended to provide occupancy predictions for obstacles with state intervals.
The proposed method can verify the safety of given trajectories (e.g.\ if they ensure collision-free fail-safe maneuver options) w.r.t.\ arbitrary safe-state formulations.
The potential for provably safe trajectory planning is shown in three evaluative scenarios.


\end{abstract}
\section{Introduction}
\label{introduction}

Over the last decades research effort about ADAS and fully automated vehicles has increased drastically in academia and the commercial sector.
The latter presenting a worrying push to early market introduction as at least two major challenges, safety and scalability, have not been solved yet~\cite{mobileye_formal_2017}.

Scalability is an issue, because most state-of-the-art approaches require sensors and processing power that is not, nor will likely be available for prices that allow mass sales in the near future~\cite{mobileye_formal_2017}.

Provable safety on the other hand is a must with regard to accountability, user acceptance and in consequence as a prerequisite for legal permission.
As extensively explained in~\cite{mobileye_formal_2017}, provable safety is of special interest when applying machine learning approaches as they often lack formal validation methods.
Therefore, they introduce the notion of blame, define a concept of Responsibility Sensitive Safety (RSS) and explain how to develop automated vehicles that provably fulfill RSS.
Another promising and meanwhile complementary approach to safety verification is the computation of reachable sets of obstacles (reachable states limited by physics and traffic law)~\cite{althoff_set-based_2016}.
These are used to prove if the ego trajectory allows a fail-safe maneuver in the next planning step and, therefore, is safe in itself.
They formally introduce the method for sets of initial obstacle states, but de facto developed only over-approximations for exactly known initial obstacle states.

\begin{figure}[!t]
  \includegraphics[width=\columnwidth]{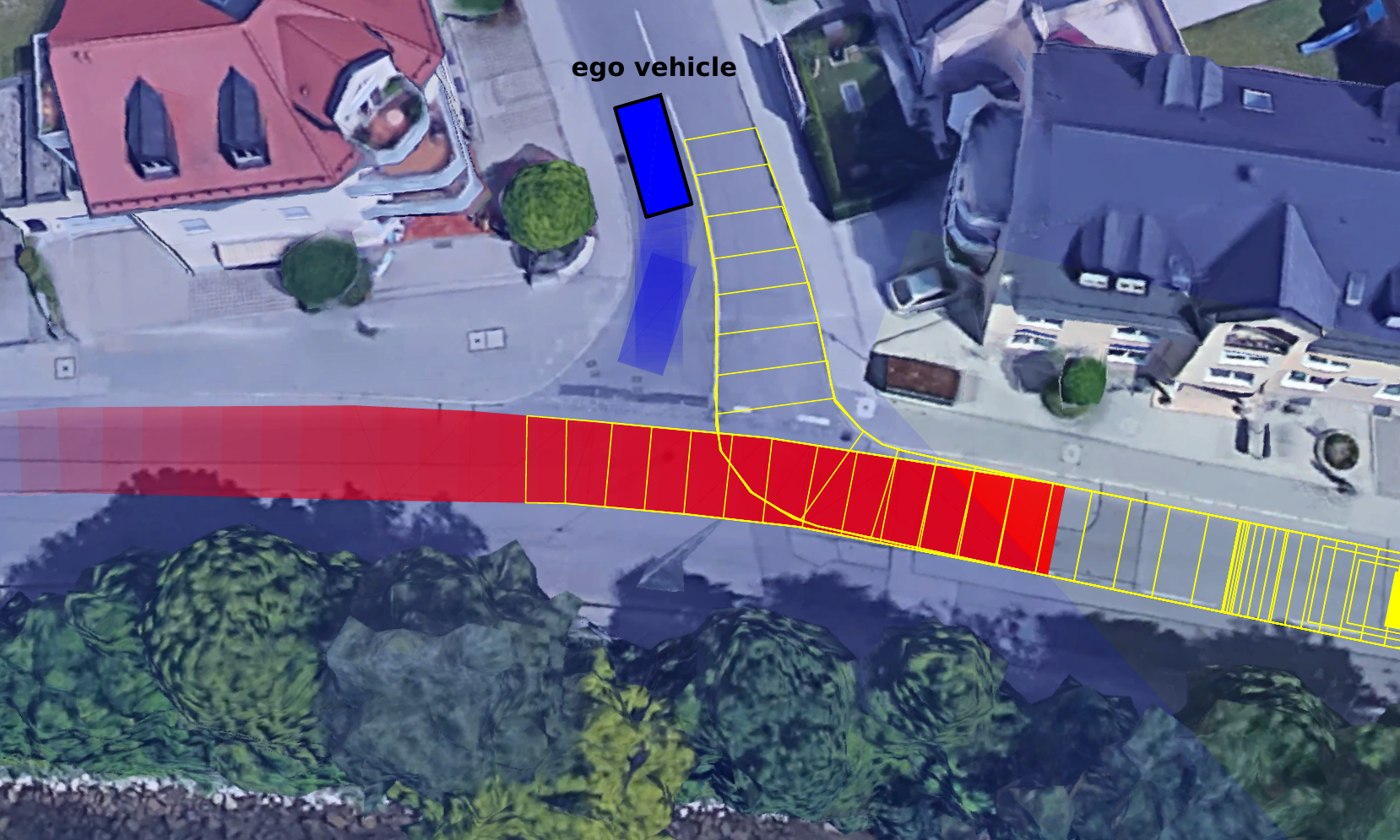}
  \caption{Dangerous intersection with occlusions.
    Yellow: The obstacle is occluded by a container, such that it and its reachable set is not known to the ego vehicle.
    Red: We can over-approximate the reachable set of possibly hidden vehicles and adapt the ego trajectory (transparent blue) to still guarantee safety.
    Imagery \copyright~2018 Google, Map data \copyright~2018 GeoBasis-DE/BKG.%
    }\label{fig:eye_catcher}
\end{figure}

In reality, we don't know exact obstacle states, because we encounter a variety of uncertainties and incomplete environment knowledge.
The former arise in all stages of an ADAS pipeline, from measurement noise and sensor limitations, over processing steps as localization, tracking, prediction and planning (due to modeling errors or unexpected situations), up to the imprecise realization of planned trajectories.
Additionally, the environment knowledge is incomplete because of limited perception range and due to occlusions from static and dynamic obstacles alike.
A typical example is illustrated in~\cref{fig:eye_catcher}.

\IEEEpubidadjcol

\section{Related Work}
\label{sec:related_work}

A lot of research has been done to approach the problem of risk assessment.
A well-arranged survey is given in~\cite{lefevre_survey_2014}.

Most of the methods presented try to develop behavior models and then
check for collisions under the assumption of those models or detect deviations from these models.
Either way such risk assessment approaches assume
that all possible maneuvers in a given scenario can be modeled
or that unexpected situations can be detected reliably.
Additionally, many do not consider the incompleteness of an environment model at all.

Recently focus on occlusion-aware%
\footnote{We will use \enquote{occlusion} for any area outside of our field of view. This includes areas occluded by obstacles, but also areas outside the sensor range.}
risk assessment and behavior generation has increased
\cite{hoermann_entering_2017,lee_collision_2017,chung_safe_2009,brechtel_probabilistic_2014,miller_isaac_team_2008,sadou_occlusions_2004,bouraine_passively_2014,zhan_non-conservatively_2016}.
These range from simple visibility modeling that improves the tracking of previously detected obstacles~\cite{miller_isaac_team_2008},
to sophisticated multi-layered environment models~\cite{hoermann_entering_2017}.

Some of them explicitly consider uncertainties~\cite{zhan_non-conservatively_2016,hoermann_entering_2017,brechtel_probabilistic_2014},
while others use visibility analysis to define velocity constraints~\cite{chung_safe_2009,lee_collision_2017}.
Only two publications among those prove at least passive motion safety%
\footnote{\enquote{If a collision takes place, the robot will be at rest.}~\cite{bouraine_passively_2014}}~\cite{bouraine_passively_2014} or
prove collision freedom at discrete time steps of their trajectories%
\footnote{Collisions between discrete time steps will not be detected here.}~\cite{zhan_non-conservatively_2016}.

A promising approach is given by~\cite{hoermann_entering_2017}.
They model and predict the environment with three grid map layers:
Object-based, object-free and unobservable environment.
During planning, they treat a cell as occupied as soon as one of the layers is predicted as occupied.

Still, these methods minimize the risk of collisions at most~\cite{brechtel_probabilistic_2014,chung_safe_2009,hoermann_entering_2017,lee_collision_2017}
and give no or too weak~\cite{bouraine_passively_2014,zhan_non-conservatively_2016} safety guarantees.
Some of the earliest approaches show even fundamental problems like ignoring not yet observed obstacles~\cite{miller_isaac_team_2008},
analyzing occluded areas without prediction~\cite{sadou_occlusions_2004}
or only investigating occlusions from static obstacles~\cite{chung_safe_2009}.

Concluding, all these methods lack verification of higher levels of safety\footnote{Meaning higher than passive or even passive friendly safety~\cite{macek_towards_2009}, but at least RSS~\cite{mobileye_formal_2017}.}
considering limited environment knowledge.

Vehicles with such risk assessment strategies will try to follow their intended behavior as long as no noteworthy risk is detected.
But this could still lead to situations where a collision is inevitable.

We want to promote an inverse methodology.
An automated vehicle should only follow its intended trajectory as long as it can prove that it is safe.
Let us ensure safety first and then increase comfort and traffic flow, not the other way around.
Such a development approach will lead to more conservative behavior in the first years,
but on the other hand help in earlier legal permission and to gain users confidence.

Thus, we tackle the safety verification problem in scenarios with occlusions
by over-approximating all possible states instead of engineering discrete maneuvers or maneuver classes.

Hence, our safety concept does not rely on e.g.\ the performance of an intention estimation module.
It depends only on the reliability of an ego localization\footnote{w.r.t.\ the map.}, the detection of obstacles, of occluded areas
and on a map\footnote{Featuring the road topology, geometry and traffic rules.}.

Therefore, in this work we characterize potential risk from occlusions and limited sensing (\cref{sec:main_risk_from_occlusion}),
enhance the reachable set over-approximations introduced by~\cite{althoff_set-based_2016} to serve well for these critical perception field bounds (\cref{sec:main_reachable_sets_for_intervals})
and show its potential for safe trajectory planning (\cref{sec:main_planning}) in several simulative scenarios (\cref{sec:results}).

Our main contribution, addressed in \cref{sec:main_risk_from_occlusion,sec:main_reachable_sets_for_intervals}, is twofold.
\begin{enumerate}
  \item We formalize the potential risk from occlusions and limited sensing by
        over-approximating all possible states of unobservable obstacles with state intervals.
  \item We derive reachable set over-approximations for obstacles with such initial state intervals.
\end{enumerate}

The approach is not only applicable for fully automated vehicles as in our evaluation, but also for level 1--4 ADAS~\cite{committee_taxonomy_2014}, e.g.\ collision warning systems.
Additionally, our modeling enables an easy integration of uncertainties from measurements
and can be used with any safety definition%
.
\section{Risk from Occlusions and Limited Sensing}
\label{sec:main_risk_from_occlusion}

As a necessary preparation for the following chapters, we first characterize the potential risk that results from occlusions and limited sensor range.

We model the environment mostly as in~\cite{althoff_set-based_2016}, meaning that the road geometry and topology is given as a lanelet map~\cite{bender_lanelets_2014}, a lane consists of consecutive lanelets, the ego vehicle and other (visible) obstacles have rectangular shape and the over-approximations of predicted occupancies are modeled with polygons.
The main difference in modeling is that we explicitly represent critical sensing field borders.

To do so we need a representation of the sensing field of the ego vehicle.
This can be provided by accurate sensor models or explicitly mapped from sensor data.
The mapping is straight forward for 3D range sensors, e.g.\ all range measurements (occupied or free) can be modeled as rays with specified beam angle and used directly as geometrical shapes or transferred into a visibility layer of a grid map at the cost of discretization errors.

\begin{figure}[!t]
  \includegraphics[width=\columnwidth]{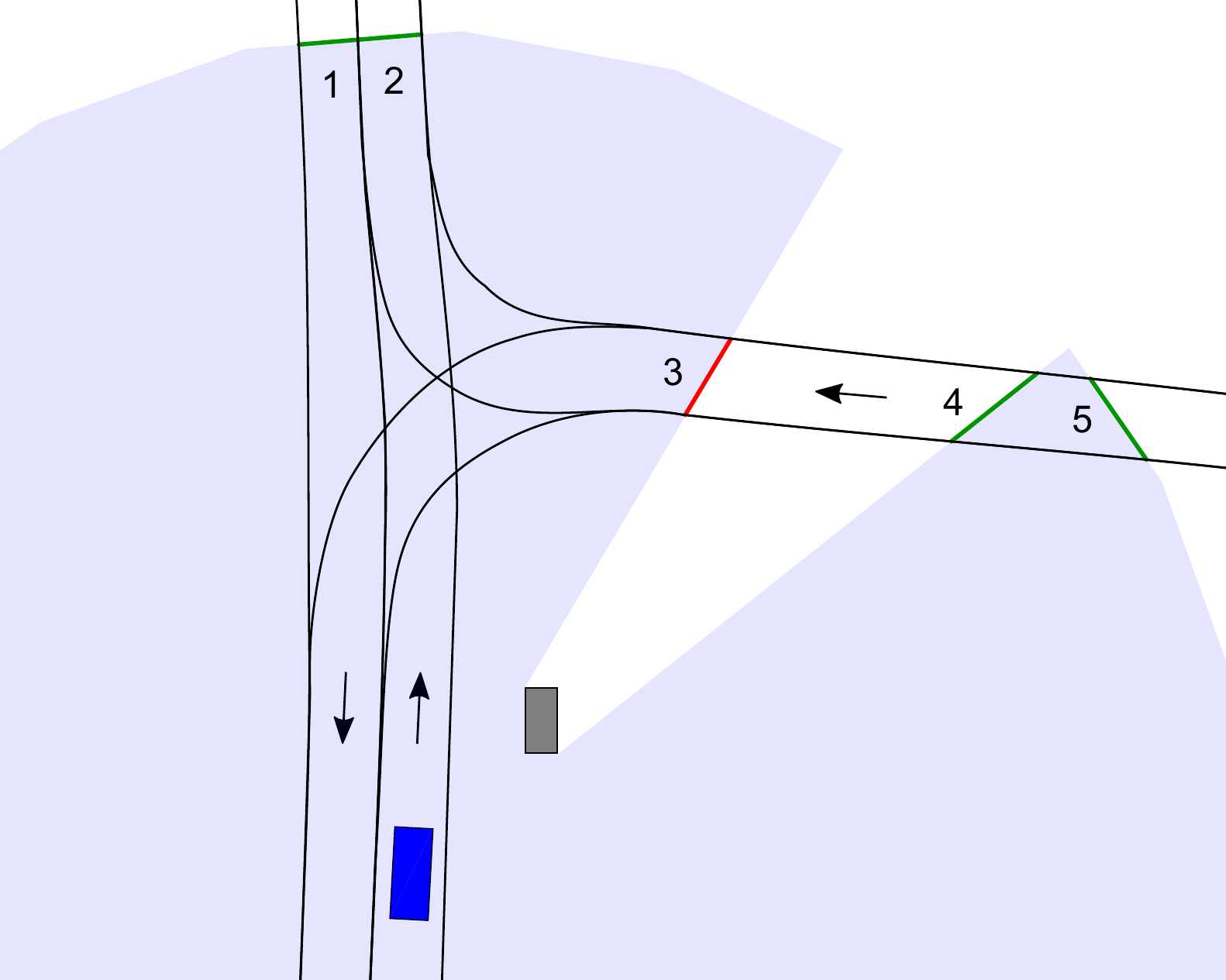}
  \caption{Critical sensing field edges.
  The one-way driving direction of each lane is visualized as an arrow.
  Relevant edges are red, irrelevant green.
  Edge~1 is not relevant as it is on a lane that does never cross or lead to lanelets we travel.
  Edge~2 is not relevant as its travel direction leads outside of the sensing field.
  Edge~3 is relevant as its lane leads to and crosses the ego lane.
  Edge~4 and 5 are not relevant as their risk is already covered by edge~3.}\label{fig:critical_edges}
\end{figure}

In our simulative evaluation we use a simple visibility model, assuming a $\ang{360}$ range sensor with $\SI{50}{\meter}$ viewing range mounted on top of the vehicle center using a direct geometrical representation, see the light blue filled area in \cref{fig:critical_edges}.

The borders of the sensing field can then be extracted and intersected with all lanelets.
Each of these intersections generates at least one border segment and can be classified into relevant and irrelevant sections.
For computational efficiency we assume the border to be modeled as a polygon, thus the border sections consist of line segments.

The classification of relevant sections is not self-evident, but can be reduced to false positives only, which are not critical.
In general only sections that could hide obstacles with right of way need to be examined.
Falsely considering non relevant sections as relevant comes at an additional performance cost, but does not affect the safety verification result.
That is because the occupancy of corresponding potential obstacles does not intersect with the current or any of the coming ego lanelets.

Employed on the lanelet representation, relevant sections can be classified as non relevant without false-positives in these incomplete cases:

\begin{itemize}
  \item Border segments that do not intersect any lanelet which can lead to or cross any of the lanelets the ego vehicle is currently or will be traveling.
  \item Segments on lanelets without right of way.
  \item Segments in the ego lane that are behind the ego vehicle.\footnote{
        This can be motivated by the concept of blame~\cite{mobileye_formal_2017} or similar, as the blame of an accident is on the rear car as long as the ego did not cut in the others' lane with an unsafe longitudinal distance.}
  \item Edges that lead outside of the sensing field\footnote{w.r.t.\ the driving direction.}.
  \item For multiple segments on the same side\footnote{\label{ft:intersection}w.r.t.\ intersection.} of the same lane only the foremost segment\cref{ft:intersection} is relevant.
\end{itemize}

See \cref{fig:critical_edges} for a descriptive example of critical sensing field edges.

Having identified the critical sensing field edges, the key question arises:
What is the potential risk that results from these?
The illustrative answer is that each occluded area might contain at least one obstacle with unknown state.
A naive approach of simply spawning countless virtual obstacles with randomly chosen possible states and predict their occupancy is not an option as a stochastic approach will not enable us to verify planned trajectories with reasonable computational cost.
But even though the state is unknown, we can over-approximate possible states as bounded sets, such that other states are either physically impossible or would not lead to an accident of our blame (as for tremendous speeding of the obstacle for example).
Therefore, we define one virtual obstacle for each critical edge $e$ with the following state set,
described as intervals in orientation $\psi_e(0)$ and velocity $v_e(0)$ and a line segment in the initial position $\bm{s}_e(0)$ defined by the two edge vertices $\bm{s}_1$ and $\bm{s}_2$.
We use the following notation:

\begin{subequations}\label{eq:x_edge}
  \begin{align}
    \bm{s}_e(0) &\in \left[ \begin{pmatrix} s_{1,x} \\ s_{1,y} \end{pmatrix}, \begin{pmatrix} s_{2,x} \\ s_{2,y} \end{pmatrix} \right]\\
    \psi_e(0) &\in [\psi_{\min}, \psi_{\max}] \\
    v_e(0) &\in [v_{\min}, v_{\max}]
  \end{align}
\end{subequations}

In the following we will enhance the reachable set approximations introduced by~\cite{althoff_set-based_2016} and released as source code~\cite{koschi_spot_2017} for such initial state sets.

\section{Reachable Sets for Intervals of Initial State}
\label{sec:main_reachable_sets_for_intervals}

In this section we derive reachable set over-approximations for obstacles with initial state intervals~\cref{eq:x_edge}.
These are based on the $M_1$ and $M_2$ over-approximations from~\cite{althoff_set-based_2016}.
$M_1$ describes the physically reachable area based on Kamm's circle, limiting the possible absolute acceleration and prohibiting driving backwards as an assumption.
To incorporate a lane-following property $M_2$ describes the longitudinally reachable area with maximum velocity and maximum engine power.
The maximum velocity should be set to a realistic value which represents expectable speeding (e.g.\ $\SI{110}{\percent}$ of the speed limit), the maximum engine power can be set to infinity.

The following subsections expand the $M_1$ and $M_2$ formulas step by step to intervals of initial state.
We keep the original notation wherever feasible, but redefine some variables to keep a clean notation, free of avoidable indices and accents.

\subsection{Acceleration-Based Occupancy $M_1$}
\label{subsec:main_reachable_sets_m1}

For simplicity, we assume the initial obstacle state to be represented in local coordinates, w.l.o.g.:

\begin{subequations}\label{eq:x_local}
  \begin{align}
    \bm{s}(0) &\in \left[ \begin{pmatrix} 0 \\ 0 \end{pmatrix} , \begin{pmatrix} \overline{s_x} \\ \overline{s_y} \end{pmatrix} \right]\\
    \psi_e(0) &\in [-\psi_{\max}, \psi_{\max}]\\
    v(0) &\in [\underline{v}, \overline{v}]
  \end{align}
\end{subequations}

We use a formulation of Kamm's circle with center $\bm{c}(t)$ and radius $r(t)$ and the function $\bm{b}(t)$ bounding that circle over time from~\cite{althoff_set-based_2016}:

\begin{subequations}\label{eq:kamms_circle}
  \begin{align}
    \bm{c}(t) &= \begin{pmatrix} s_x(0) \\ s_y(0) \end{pmatrix} + \begin{pmatrix} v_x(0) \\ v_y(0) \end{pmatrix} t\\
    r(t) &= \tfrac{1}{2}a_{\max}t^2\\
    b_x(t) &= v_0t-\frac{a^2_{\max}t^3}{2v_0}\\
    b_y(t) &= \sqrt{\tfrac{1}{4}a^2_{\max}t^4-\left(\frac{a^2_{\max}t^3}{2v_0}\right)^2}
  \end{align}
\end{subequations}

Please see \cref{fig:occ_vel} for a graphical representation.

\subsubsection{Interval of Initial Velocities}
\label{subsec:main_reachable_sets_m1_vel}

\begin{figure*}
  \centering
  \begin{subfigure}[!t]{0.4\textwidth}
    \includegraphics[width=\columnwidth]{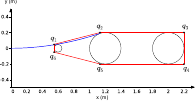}
    \caption{}\label{fig:occ_vel}
  \end{subfigure}
  \begin{subfigure}[!t]{0.3\textwidth}
    \includegraphics[width=\columnwidth]{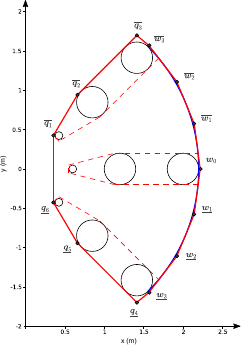}
    \caption{}\label{fig:occ_vel_xi}
  \end{subfigure}%
  \begin{subfigure}[!t]{0.25\textwidth}
    \includegraphics[width=\columnwidth]{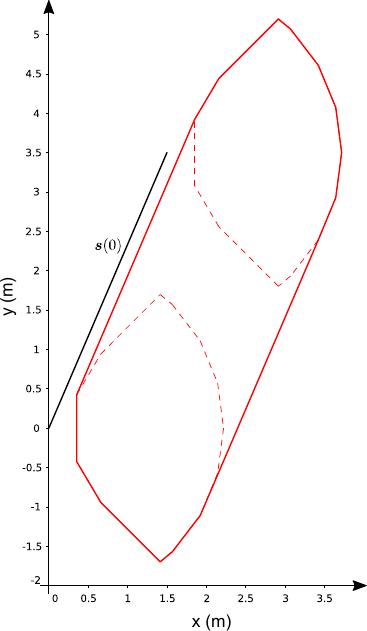}
    \caption{}\label{fig:occ_vel_xi_pos}
  \end{subfigure}

  \caption{Occupancy over-approximations for initial state intervals.
    $t_1=\SI{0.1}{\second}$, $t_2=\SI{0.2}{\second}$, $a_{\max}=\SI{10}{\mpss}$
    \subref{fig:occ_vel}
      Initial velocity as interval: $\underline{v}=\SI{6}{\mps}$, $\overline{v}=\SI{10}{\mps}$.
      Black: Kamm's circle.
      Blue: $\bm{b}(t)$ bounding Kamm's circle.
    \subref{fig:occ_vel_xi}
      Additionally orientation as interval: $\psi_{\max}=\ang{45}$, $n=3$.
      Blue: circle sector given through rotation of $\bm{p}_{long}$.
      $\underline{\bm{w}_j}, \bm{w}_0, \overline{\bm{w}_j}$: polygon over-approximating that circle sector.
      $\overline{\bm{q}_1}, \overline{\bm{q}_2}, \overline{\bm{q}_3}$,
      $\underline{\bm{q}_4}, \underline{\bm{q}_5}, \underline{\bm{q}_6}$: counterclockwise and clockwise occupancy border.
    \subref{fig:occ_vel_xi_pos}
      Additionally initial position as line segment: $\overline{s_x}=\SI{1.5}{\meter}$, $\overline{s_y}=\SI{3.5}{\meter}$.
  }\label{fig:occ}
\end{figure*}

With an interval in the initial velocity $v_0 \in [\underline{v}, \overline{v}]$, but known orientation $\psi(0)=0$ and $\bm{s}(0)=(0,0)^T$, we can define

\begin{align}\label{eq:c_interval}
  \underline{c}(t) &= c(t, \underline{v}), \quad
  \overline{c}(t) = c(t, \overline{v})
\end{align}

and $b_x, b_y$ likewise.
The occupancy of the obstacle for a time period of $\tau_k = [t_k, t_{k+1}]$ can then be over-approximated by the polygon spanned by the points $\bm{q}_1, \dots, \bm{q}_6$:

\begin{subequations}\label{eq:occ_vel}
  \begin{align}
    \bm{q}_1 &= \left(\underline{c_x}(t_k)-r(t_k),\ r(t_k)\right)^T\\
    \bm{q}_2 &= \left(\underline{b_x}(t_{k+1}),\ r(t_{k+1})\right)^T\\
    \bm{q}_3 &= \left(\overline{c_x}(t_{k+1})+r(t_{k+1}),\ r(t_{k+1})\right)^T\\
    \bm{q}_4 &= \left(\overline{c_x}(t_{k+1})+r(t_{k+1}),\ -r(t_{k+1})\right)^T\\
    \bm{q}_5 &= \left(\underline{b_x}(t_{k+1}),\ -r(t_{k+1})\right)^T\\
    \bm{q}_6 &= \left(\underline{c_x}(t_k)-r(t_k),\ -r(t_k)\right)^T
  \end{align}
\end{subequations}

as visualized in \cref{fig:occ_vel}.

The left part is equivalent to the original $\mathcal{O}_1(\tau_k,\underline{v})$, but vertices $\bm{q}_3$, $\bm{q}_4$ are computed using $\overline{v}$.
This occupancy encloses all $v_i \in [\underline{v}, \overline{v}]$ as each circle $C_{k+1}(v_i)$ has the same radius $r(t_{k+1})$, and center $c_y=0, {c_x}(t_{k+1}) \in [\underline{c_x}(t_{k+1}), \overline{c_x}(t_{k+1})]$.
Therefore, it is enclosed by the polygon $P(\bm{q}_1, \bm{q}_2, \bm{q}_3, \bm{q}_4, \bm{q}_5, \bm{q}_6)$ spanned from $\underline{C_{k}}$, $\underline{C_{k+1}}$ and $\overline{C_{k+1}}$.
Consequently, this polygon encloses all $C_{t}(v_i)$ with $t \in [t_k, t_{k+1}]$, proving that this polygon is an over-approximation of all reachable states for an obstacle with interval velocities.

\subsubsection{Interval of Initial Orientations}
\label{subsec:main_reachable_sets_m1_phi}

With additionally an initial orientation interval of $\psi(0) \in [-\psi_{\max},\psi_{\max}]$ the whole occupancy $P(\bm{q}_1, \bm{q}_2, \bm{q}_3, \bm{q}_4, \bm{q}_5, \bm{q}_6)$ rotates around the origin depending on the real initial orientation of the obstacle.
We can again over-approximate this set.
The borders of the set can be derived by rotating $\bm{q}_1, \bm{q}_2, \bm{q}_3$ counterclockwise to $\overline{\bm{q}_1}, \overline{\bm{q}_2}, \overline{\bm{q}_3}$ and $\bm{q}_4, \bm{q}_5, \bm{q}_6$ clockwise to $\underline{\bm{q}_4}, \underline{\bm{q}_5}, \underline{\bm{q}_6}$ by $\psi_{\max}$
and over-approximating the circle given through the rotation of the furthest longitudinal point $\bm{p}_{long} = (\overline{c_x}(k+1) + r(t_{k+1}),\ 0)^T$ with

\begin{subequations}\label{eq:eq_circle_approx}
  \begin{align}
    \bm{w}_0 &= \left(\frac{\overline{c_x}(t_{k+1})+r(t_{k+1})}{\cos\frac{\theta}{2}},\ 0\right)^T,
      \quad \theta=\frac{\psi_{\max}}{n}\\
    \overline{\bm{w}_j} &= R_{\theta_j}\bm{w}_0\\
    \underline{\bm{w}_j} &= -R_{\theta_j}\bm{w}_0
  \end{align}
\end{subequations}

for integers $j \in [1, n]$.
A reasonable approximation of the circle over $\psi_{\max}$ of around $\ang{45}$ can already be achieved with $n=3$.

\cref{fig:occ_vel_xi} illustrates the construction.
The prove that each polygon with rotation $\psi_i \in [-\psi_{\max},\psi_{\max}]$ is enclosed by the polygon

\begin{align}\label{eq:polygon_vel_xi}
  P(\overline{\bm{q}_1}, \overline{\bm{q}_2}, \overline{\bm{q}_3},
    \overline{\bm{w}_n}, \dots, \overline{\bm{w}_1}, \bm{w}_0, \underline{\bm{w}_1}, \dots, \underline{\bm{w}_n},
    \underline{\bm{q}_4}, \underline{\bm{q}_5}, \underline{\bm{q}_6})
\end{align}

follows trivially from its construction.

\subsubsection{Interval of Initial Positions}
\label{subsec:main_reachable_sets_m1_pos}

The transfer to position intervals, meaning linear interpolations between both line segment vertices,
can be realized by computing the occupancy $\underline{P}$ for $\underline{\bm{s}}(0)=(0, 0)^T$ as described in the previous subsection,
creating a duplicate $\overline{P}$ that has been translated by $\overline{\bm{s}}(0)=(s_x, s_y)^T$ and
computing the convex hull over both occupancies $\mathcal{O}_1(\tau_k)=\text{Conv}(\underline{P}, \overline{P})$.

The occupancy for each
$\bm{s}_i \in \left[\left(\begin{smallmatrix}0\\0\end{smallmatrix}\right), \left(\begin{smallmatrix}s_x\\s_y\end{smallmatrix}\right)\right]$
(each possible position on the line segment) is enclosed by the convex hull $\mathcal{O}_1(\tau_k)$ due to the linearity of translations and line segments.
Thus, $\mathcal{O}_1(\tau_k)$ is provably an over-approximation of the reachable set of an obstacle with an unknown but bounded initial state, modeled as~\cref{eq:x_local}.

The resulting over-approximation is visualized for realistic example parameters in \cref{fig:occ_vel_xi_pos}.

\subsection{Lane-Following Occupancy $M_2$}
\label{subsec:main_reachable_sets_m2}

The adaption of the original $M_2$ formulation for interval initial states follows quickly.
We use the same computation of shortest paths in lanes as~\cite{althoff_set-based_2016}, but choose the start- and end-point wisely from our interval.
To do so, we first sort both vertices of the line segment $\bm{s}_e(0)$ w.r.t.\ the longitudinal path coordinates of the corresponding lane or lanelet
and name them $\underline{\bm{s}}(0)$, $\overline{\bm{s}}(0)$ such that $\underline{\bm{s}}(0) < \overline{\bm{s}}(0)$ without loss of generality.
The start border of the lane occupancy is then given by $b_{\bm{s}tart}(\underline{\bm{s}_x}(0), \underline{\bm{s}_y}(0), \underline{v}(0))$.
The maximal traveled distance $\xi_f(t)$ can be obtained with the limited maximum speed and engine power model as in~\cite{althoff_online_2014}:

\begin{align}\label{eq:acceleration_model}
  a_{c2,long} =
    \begin{cases}
      a_{\max}\frac{v_S}{v},  & v_S<v<v_{\max} \land u_2>0,\\
      a_{\max},               & (0<v \leq v_S \lor (v > v_S \land u_2 \leq 0)\\
      0,                      & v \leq 0 \lor (v \geq v_{\max} \land u_2 \geq 0)
    \end{cases}
\end{align}

This enables us to compute the end border using the inflection point segmentation algorithm as $b_{end}(\overline{\bm{s}_x}(0), \overline{\bm{s}_y}(0), \overline{v}(0))$.

Finally, the occupancy prediction is obtained by computing and intersecting the reachable set approximations $M_1$ and $M_2$ for each critical sensing field edge in each prediction step period $\tau$ over the whole prediction horizon $[0, t_f]$.
A trajectory planner can use this prediction to verify if its intended trajectory is safe or not.

Note that in case of known initial state, where the intervals collapse to precise values $\bm{s}_e(0)=[\bm{s}_0, \bm{s}_0]$, $\psi(0)=[\psi_0, \psi_0]$, $v(0)=[v_0, v_0]$,
our formulation results in the same reachable set over-approximation as with the original method in~\cite{althoff_set-based_2016}.

\section{Trajectory Planning with Reachable Sets}
\label{sec:main_planning}

The strategy for trajectory planning has been profoundly motivated and extensively explained in~\cite{althoff_set-based_2016}.
Yet we want to briefly summarize the method and explain our proof-of-concept implementation to give a better understanding of the following evaluation.

The planning approach assumes that we start in a safe state with a reliable planning frequency of $\frac{1}{{\scriptscriptstyle\Delta}t}$.
The safety verification relies on induction.
Starting from a safe state with a verified fail-safe trajectory at hands the planner will search for a desirable trajectory that will guarantee a fail-safe maneuver choice in the next planning step too.
If it fails to find one, it switches to the fail-safe trajectory that has been found and verified in the previous planning step.
Thus, the vehicle will always follow a trajectory whose safety has been proven.

To do so a reference trajectory will be planned in each planning step $t$,
based on the current environment model.
This might also contain reasonably good predictions of all obstacles, though it is not a prerequisite.
But the better the prediction the higher the probability for a comfortable ride.
As unexpected situations become rare events, the planner will rarely be forced to switch to the fail-safe maneuver.

In order to generate safe trajectories a potential trajectory is computed.
It consists of an intended part, the first part of the reference trajectory over $[t, t + \dt]$, and a following fail-safe part over $[t + \dt, t_\text{f}]$ towards a safe state.
If we can prove that the potential trajectory does not intersect with any of the reachable set over-approximations of other obstacles, it is verified as safe.
If the verification fails, the intended trajectory will be iteratively adapted until a verifiably safe trajectory,
in the worst case the fail-safe trajectory of the previous planning step $t - \dt$, is found.

Our proof-of-concept planner initializes the reference trajectory according to the intelligent driver model~\cite{intelligent_driver_model} along the centerline of the ego lane.
If the verification fails the intended acceleration is gradually changed towards the fail-safe trajectory until the potential trajectory can be verified as safe.

The planning and prediction horizon $t_\text{f}$ is defined by the time needed to reach a safe state.
This raises two questions:
What is a safe state?
And in case we need to decide for a fail-safe maneuver, do we want to reach the safe state fast or comfortably?

The safe state itself depends on the intended maneuver.
As long as we don't want to cross or intersect with other lanes a full stop can be seen as safe state in urban areas.
But as soon as we merge into or cross the traffic of another lane we need to ensure a safe cut-in time.
Hence, a full stop is not a safe state in these cases.
For a further discussion on safe states please refer to e.g.~\cite{mobileye_formal_2017}.

Choosing the desired trajectory towards a safe state is a trade-off between the comfort during a fail-safe maneuver and the probability of having to switch to one.
Comfortable trajectories i.a.\ have low jerk and acceleration such that they need a long time to reach the safe state.
This durations then defines the required prediction horizon for safety verification.
With longer prediction horizons prediction occupancies are bigger, such that it becomes more likely that
replanning is necessary to find a trajectory that succeeds in verification.
As a consequence, if the prediction horizon is too long due to too comfortable fail-safe trajectories,
the probability of too conservative behavior (from a users' or traffic participants' perspective) will be high.
In our evaluation we used a compromise desired fail-safe deceleration of $\SI{4}{\mpss}$.
\section{Results and Evaluation}
\label{sec:results}

We evaluated our trajectory validation method on three critical urban scenarios with occlusion from static or dynamic obstacles
and show its usability for safe trajectory planning.
All scenarios represent or are inspired by real intersections in the city of Karlsruhe and a small town Fürstenfeldbruck.
They are based on the \textit{commonroad} scenarios \textit{DEU\_Ffb\nobreakdash1\_1\_T\nobreakdash1} and \textit{DEU\_Ffb\nobreakdash2\_1\_T\nobreakdash1}~\cite{althoff_commonroad_2017}.
Our modified versions have been contributed to the \textit{commonroad} benchmark as scenarios \textit{DEU\_Ffb\nobreakdash1\_4}, \textit{DEU\_Ffb\nobreakdash1\_5\_T\nobreakdash1} and \textit{DEU\_Ffb\nobreakdash2\_3\_T\nobreakdash1}.

We assume perfect perception from a $\ang{360}$ range sensor with $\SI{50}{\meter}$ viewing range mounted on top of the vehicle center.
The desired ego velocity is set to $\SI{9}{\mps}$ with a maximum comfortable acceleration of $\SI{2}{\mpss}$ and maximum possible acceleration of $\SI{8}{\mpss}$, a realistic value for dry asphalt~\cite{wallman_friction_2001}.
For an easier understanding of the effects of occupancies, we do not adapt the desired velocity based on curvature.
The prediction and planning horizon is defined by a desired fail-safe deceleration of $\SI{4}{\mpss}$,
a compromise between comfortable fail-safe trajectories and keeping the desired driving speed.
Other occupancy prediction parameters are $t_\text{f} = \SI{9}{\mps}/\SI{4}{\mpss} = \SI{2,25}{\second}$, $\dt=\SI{0.1}{\second}$, $a_{\max}=\SI{10}{\mpss}$, $\underline{v}=\SI{0}{\mps}$, $\overline{v}=\num{1.1} \cdot v_{\text{lim}}$, $\psi_{\max}=\ang{22.5}$, $n=3$.

\begin{figure*}[!t]
  \centering
  \begin{subfigure}[!t]{0.32\textwidth}
    \includegraphics[width=\columnwidth]{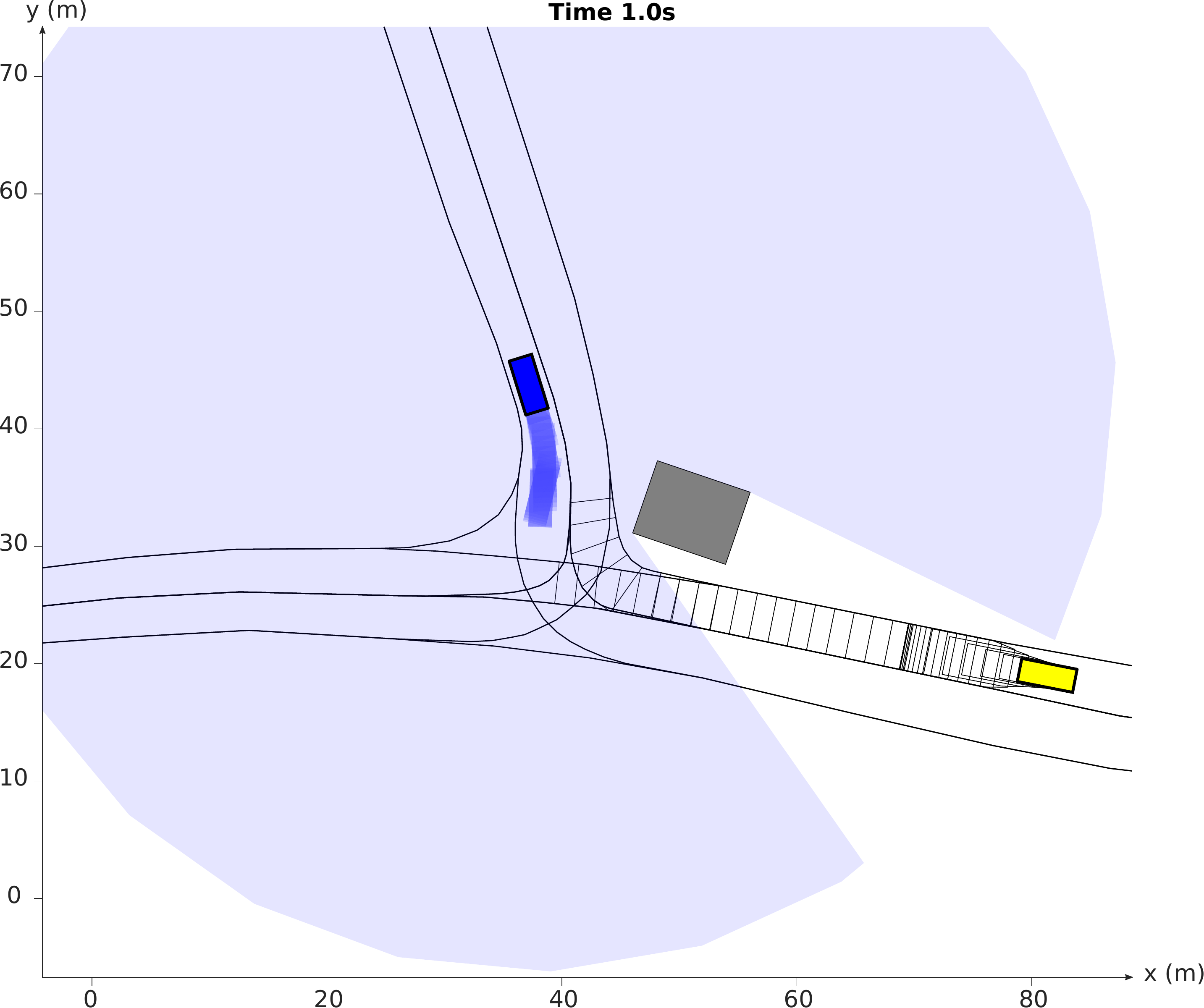}
  \end{subfigure}
  \vspace{.1em}
  \begin{subfigure}[!t]{0.32\textwidth}
    \includegraphics[width=\columnwidth]{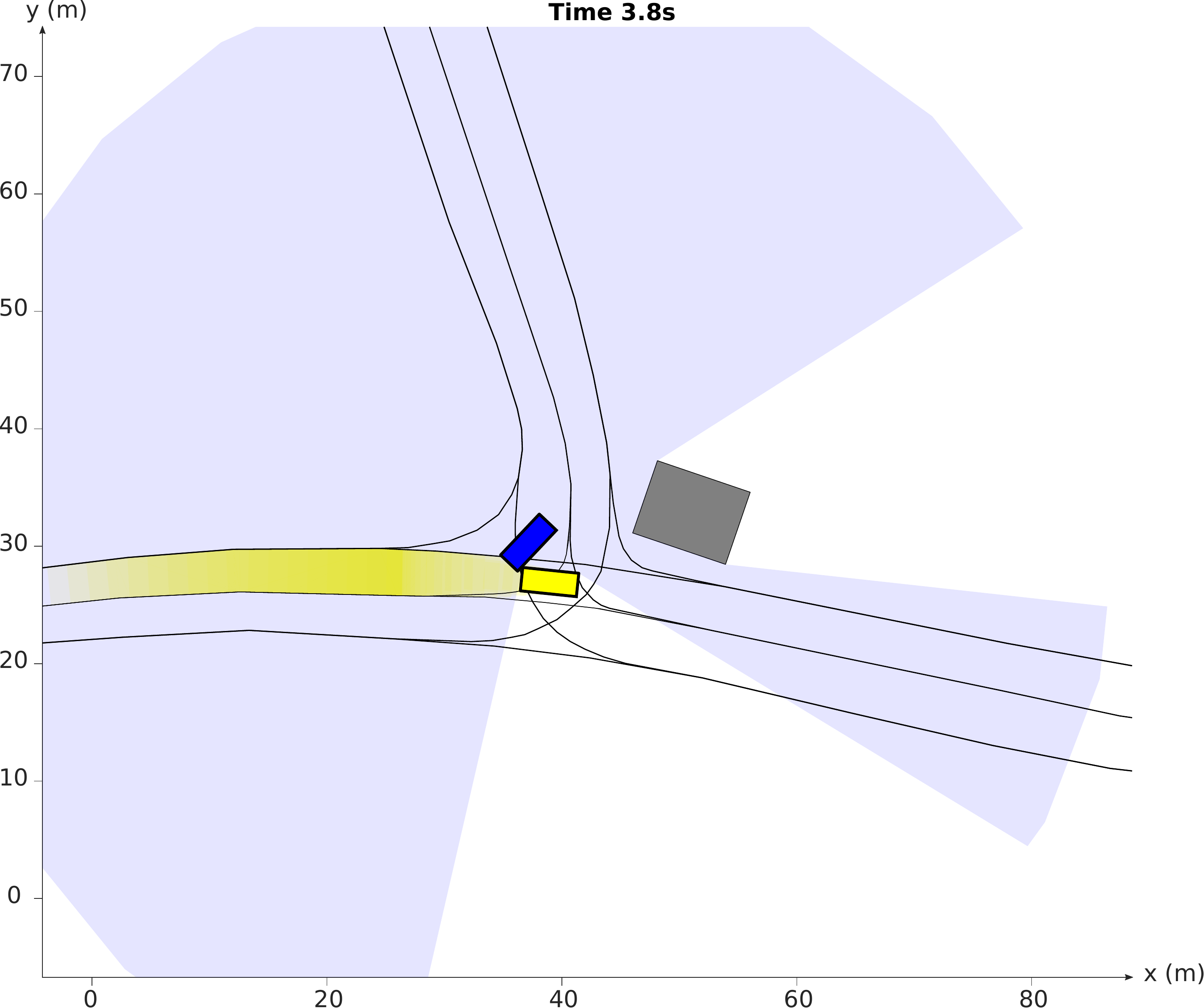}
  \end{subfigure}
  \vspace{.1em}
  \begin{subfigure}[!t]{0.33\textwidth}
    \includegraphics[width=\columnwidth]{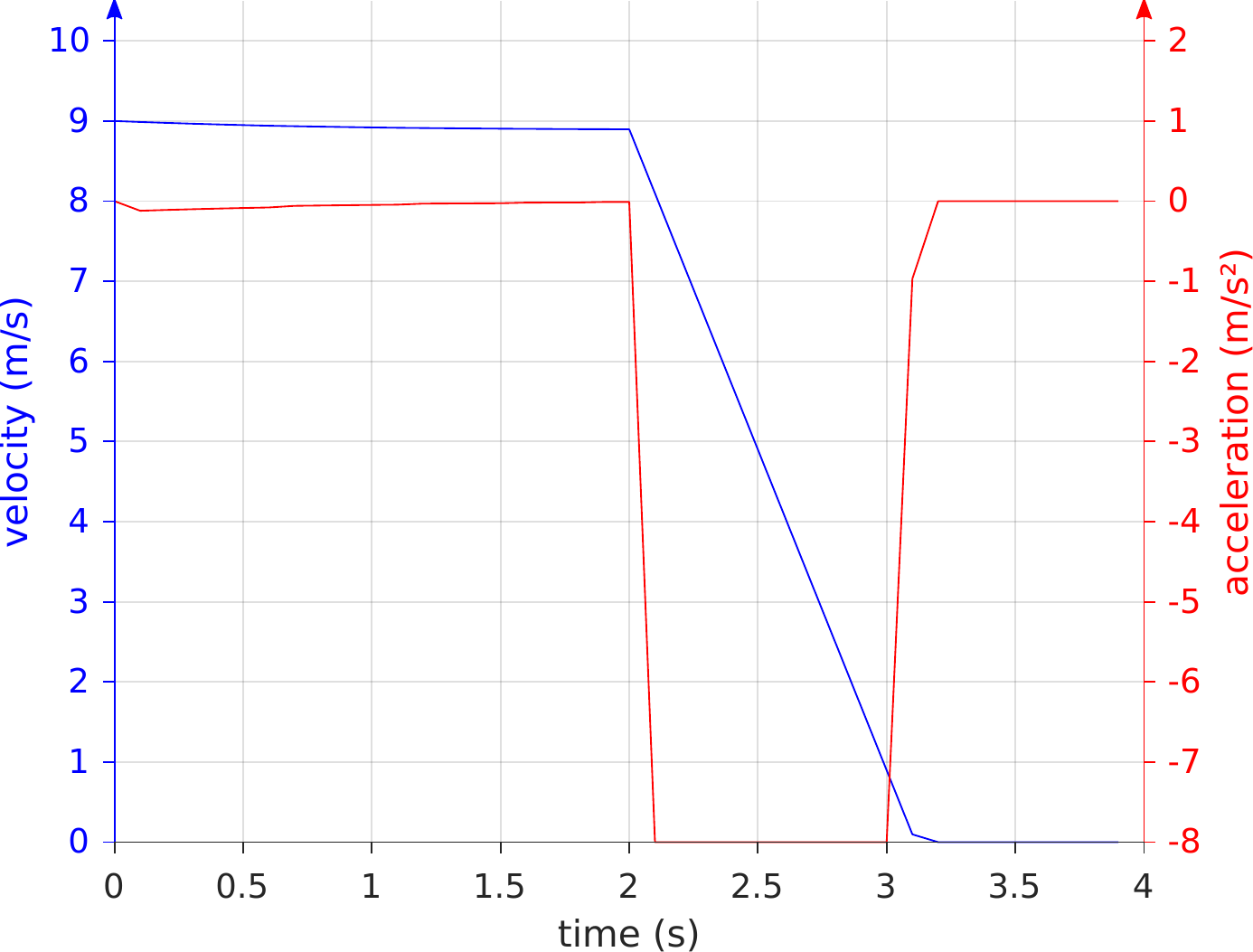}
  \end{subfigure}
  \vspace{.1em}
  \begin{subfigure}[!t]{0.32\textwidth}
    \includegraphics[width=\columnwidth]{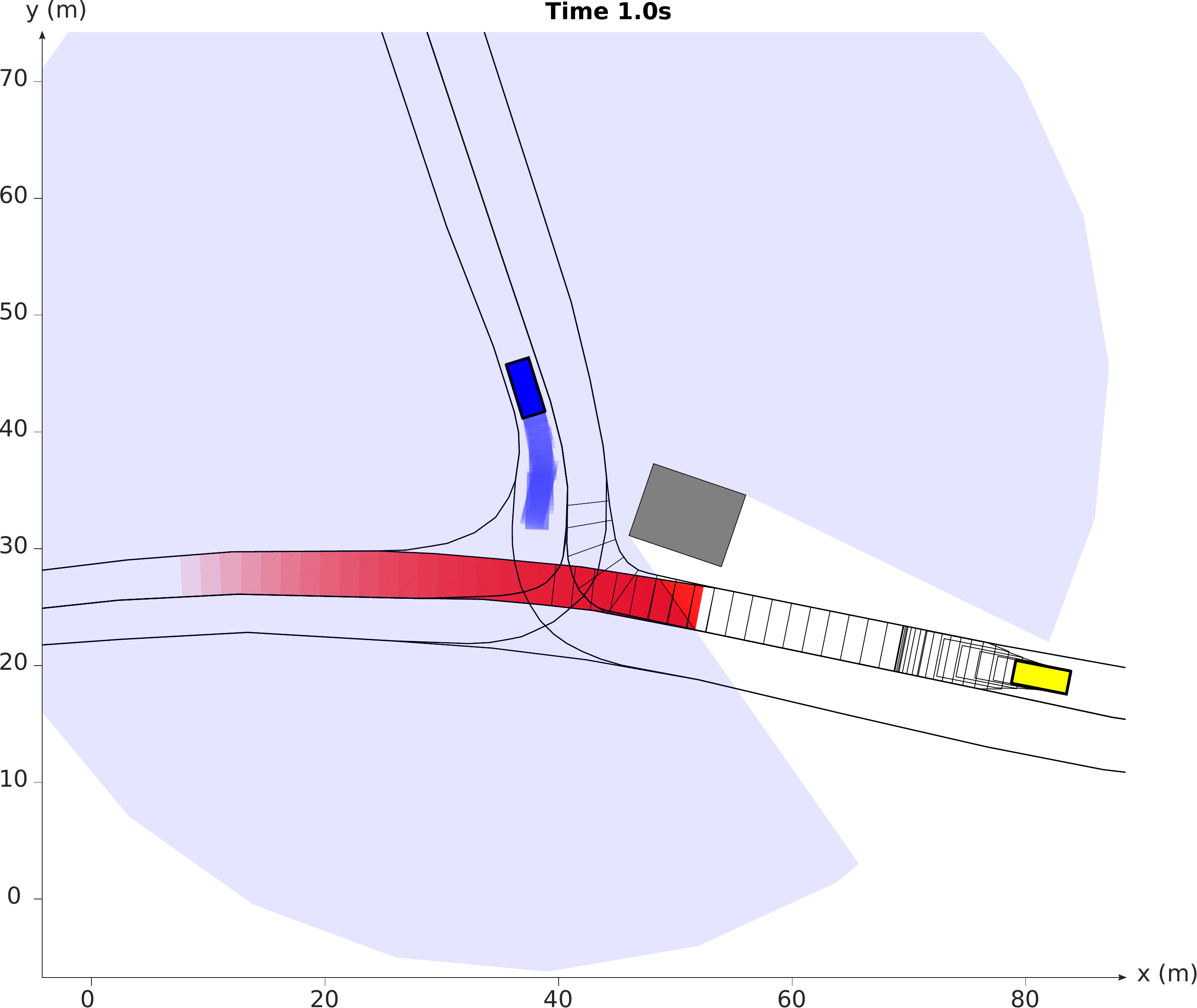}
  \end{subfigure}
  \vspace{.1em}
  \begin{subfigure}[!t]{0.32\textwidth}
    \includegraphics[width=\columnwidth]{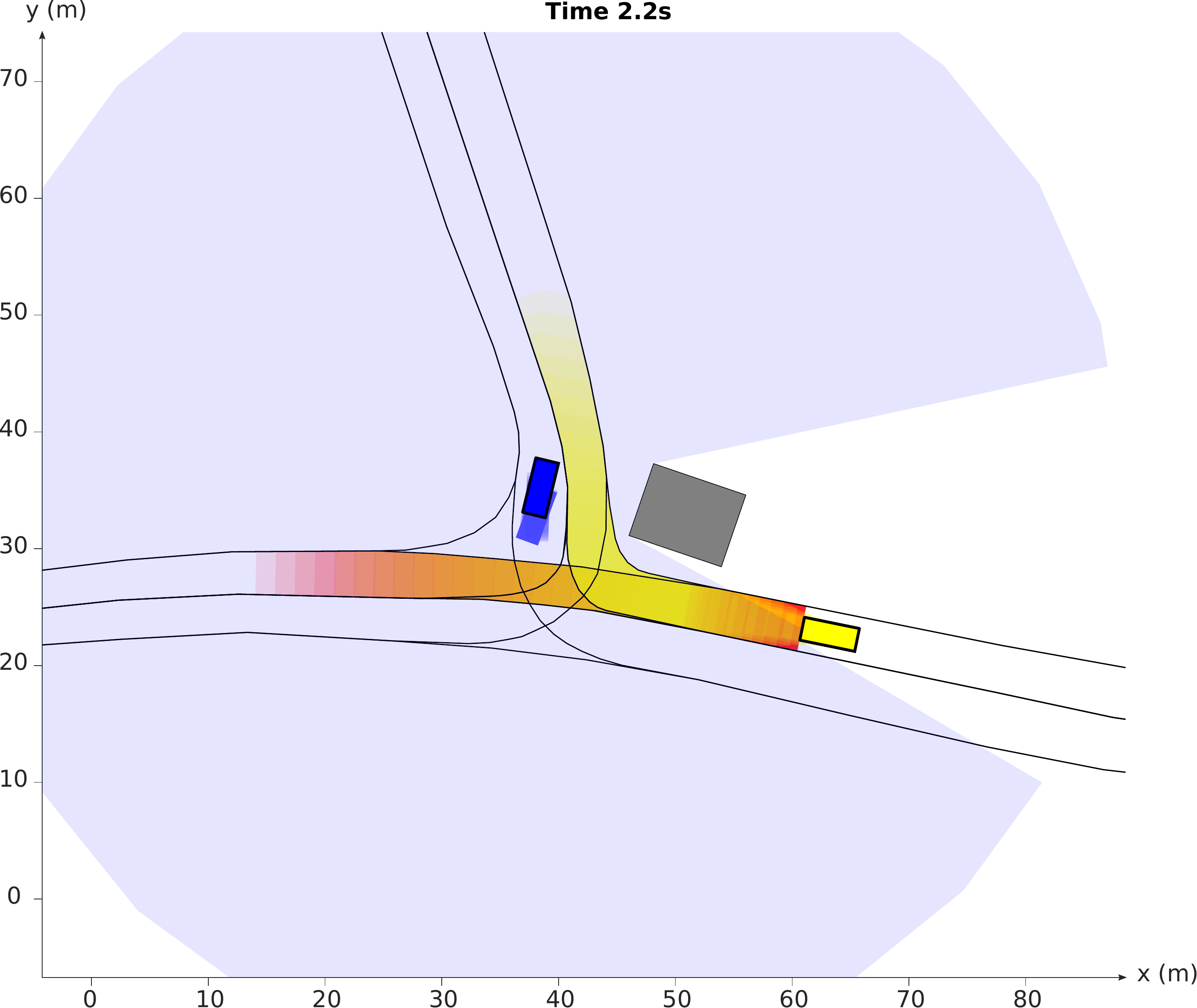}
  \end{subfigure}
  \vspace{.1em}
  \begin{subfigure}[!t]{0.33\textwidth}
    \includegraphics[width=\columnwidth]{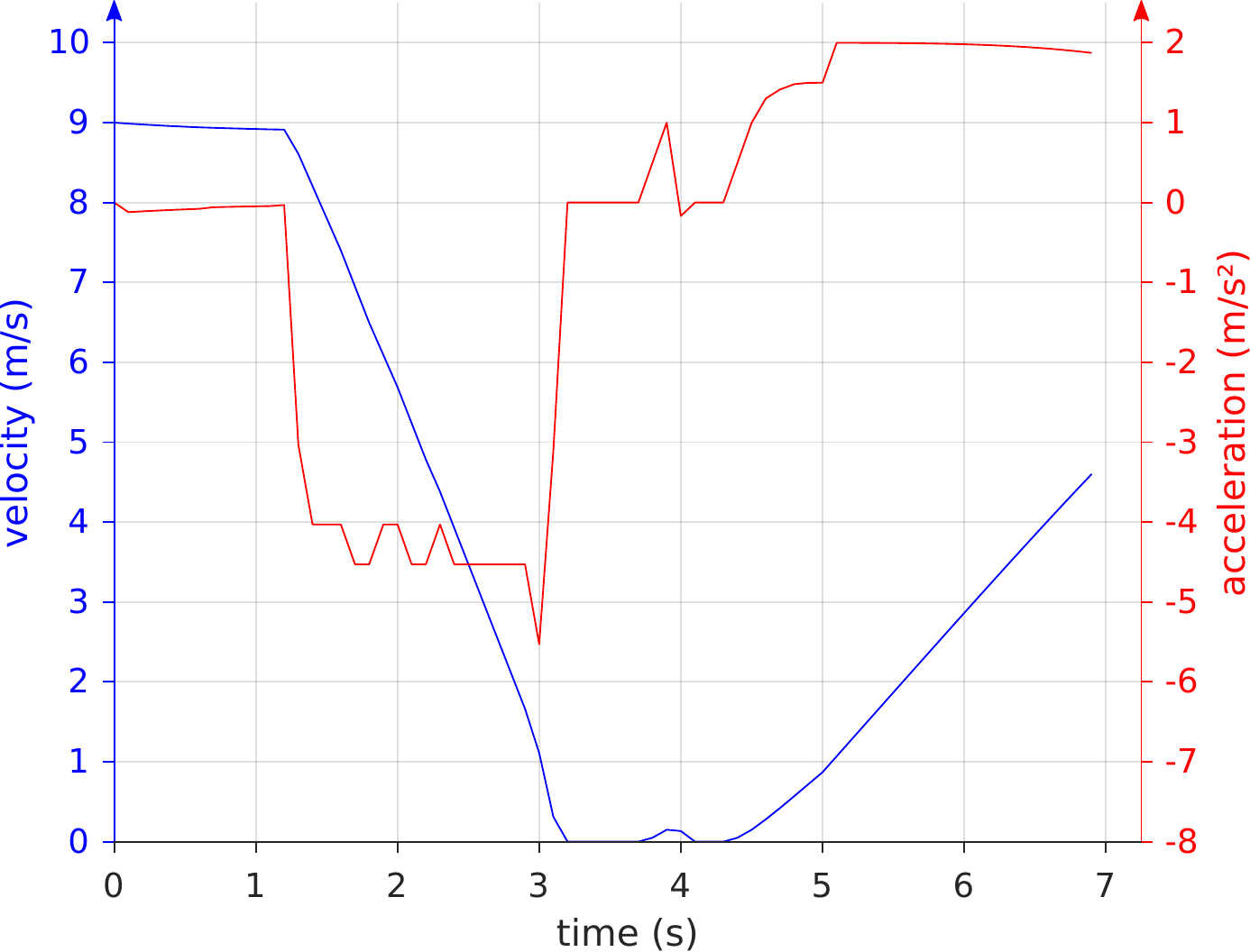}
  \end{subfigure}
  \vspace{.1em}
  \begin{subfigure}[!t]{0.32\textwidth}
    \includegraphics[width=\columnwidth]{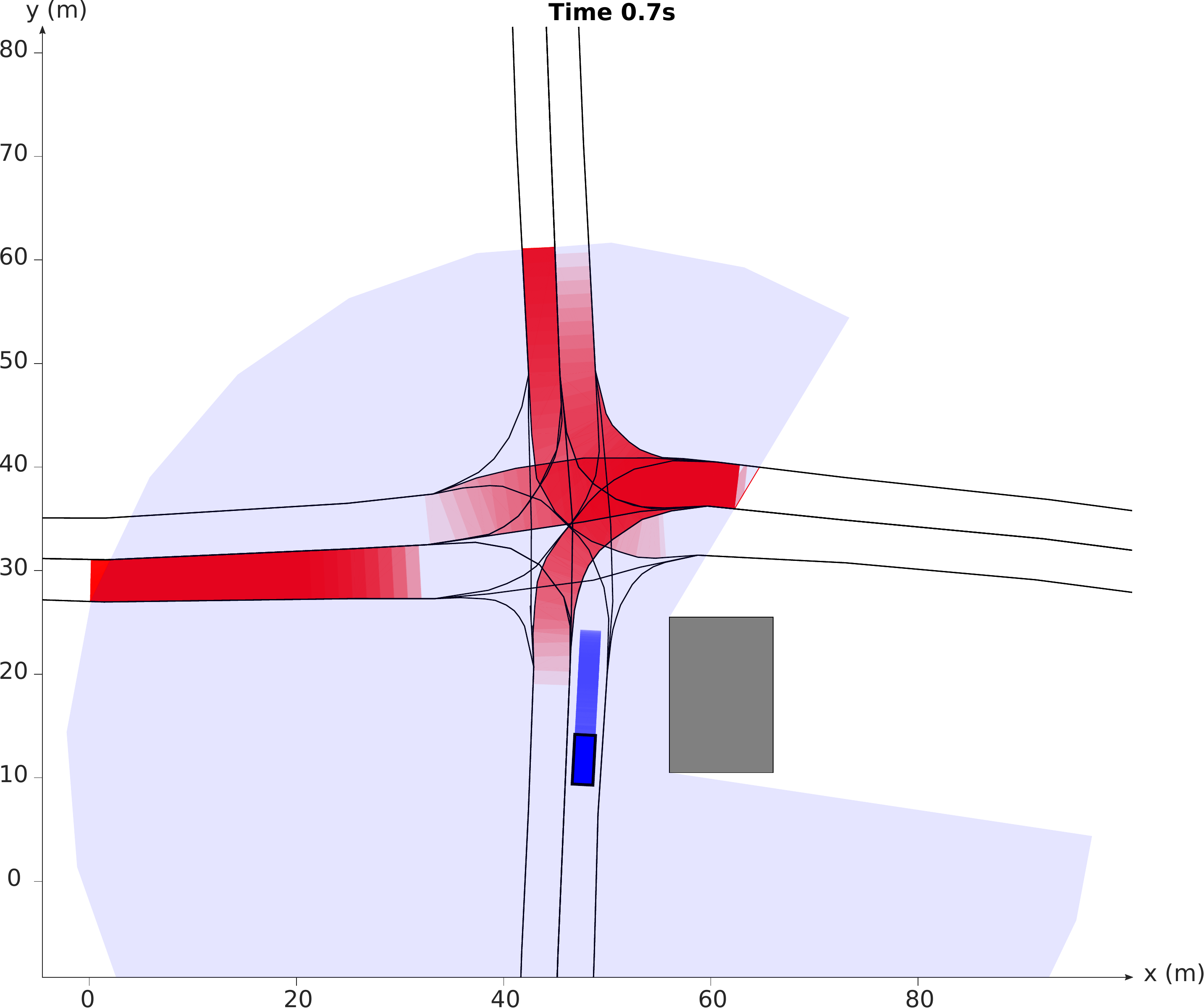}
  \end{subfigure}
  \vspace{.1em}
  \begin{subfigure}[!t]{0.32\textwidth}
    \includegraphics[width=\columnwidth]{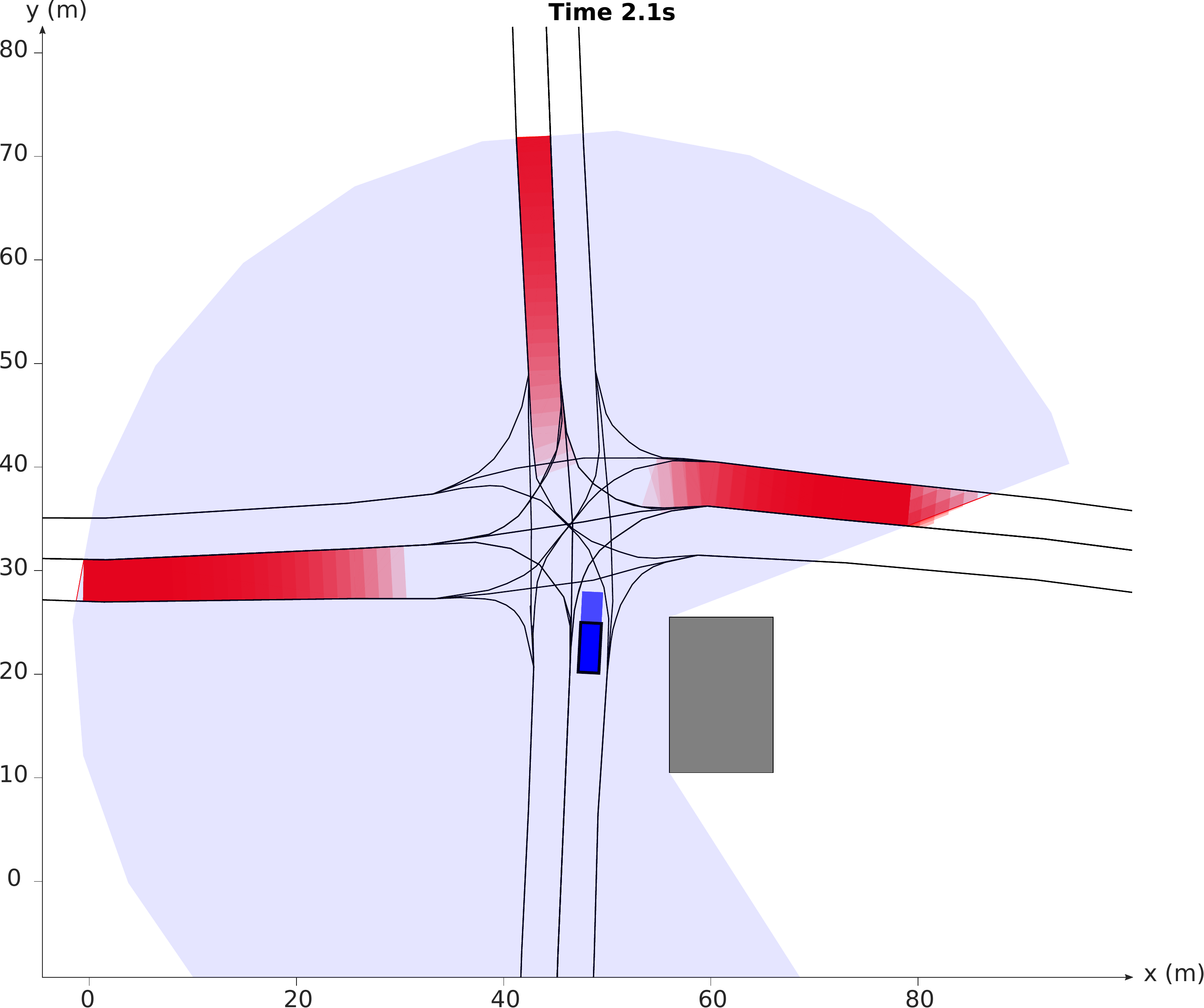}
  \end{subfigure}
  \vspace{.1em}
  \begin{subfigure}[!t]{0.33\textwidth}
    \includegraphics[width=\columnwidth]{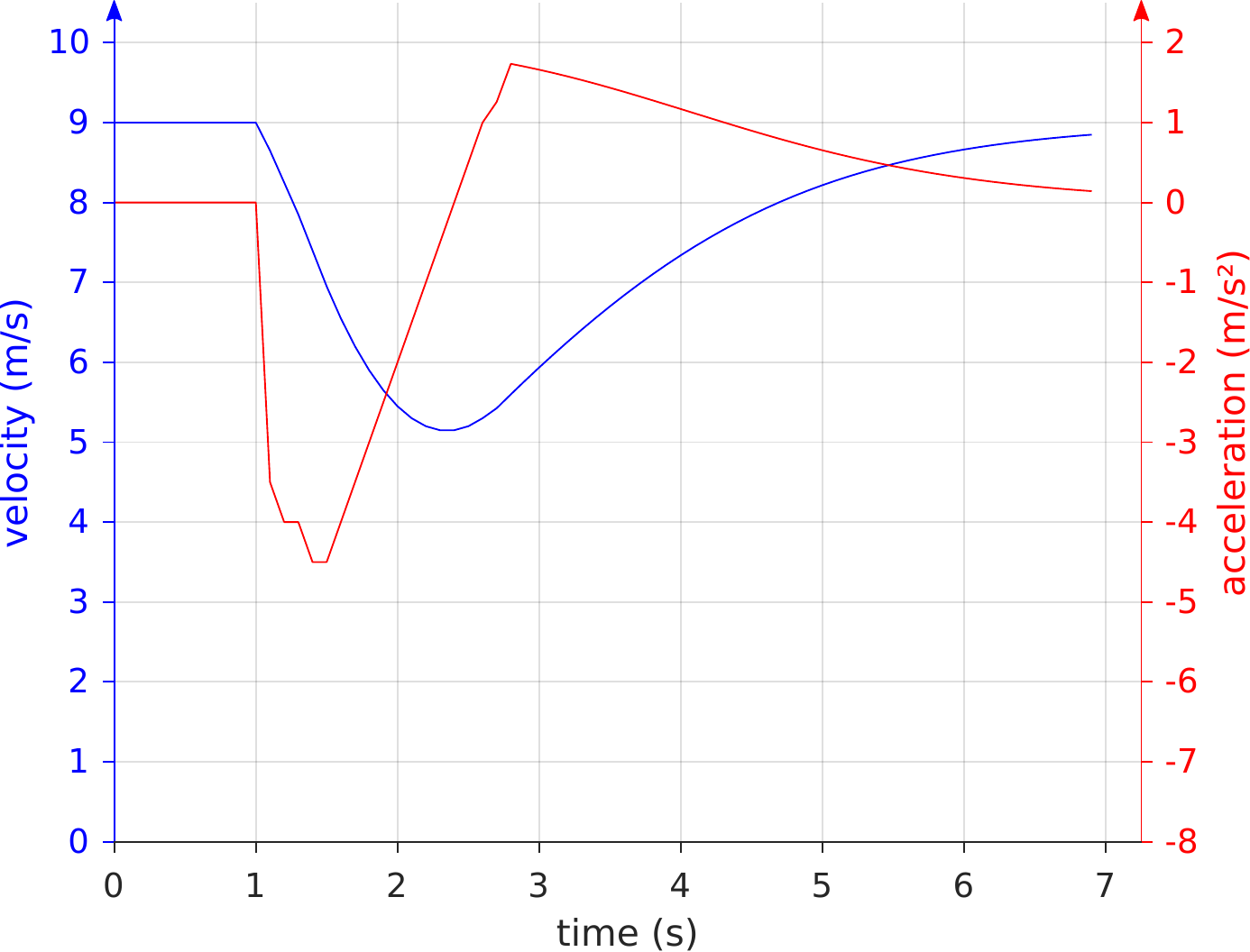}
  \end{subfigure}
  \vspace{.1em}
  \begin{subfigure}[!t]{0.32\textwidth}
    \includegraphics[width=\columnwidth]{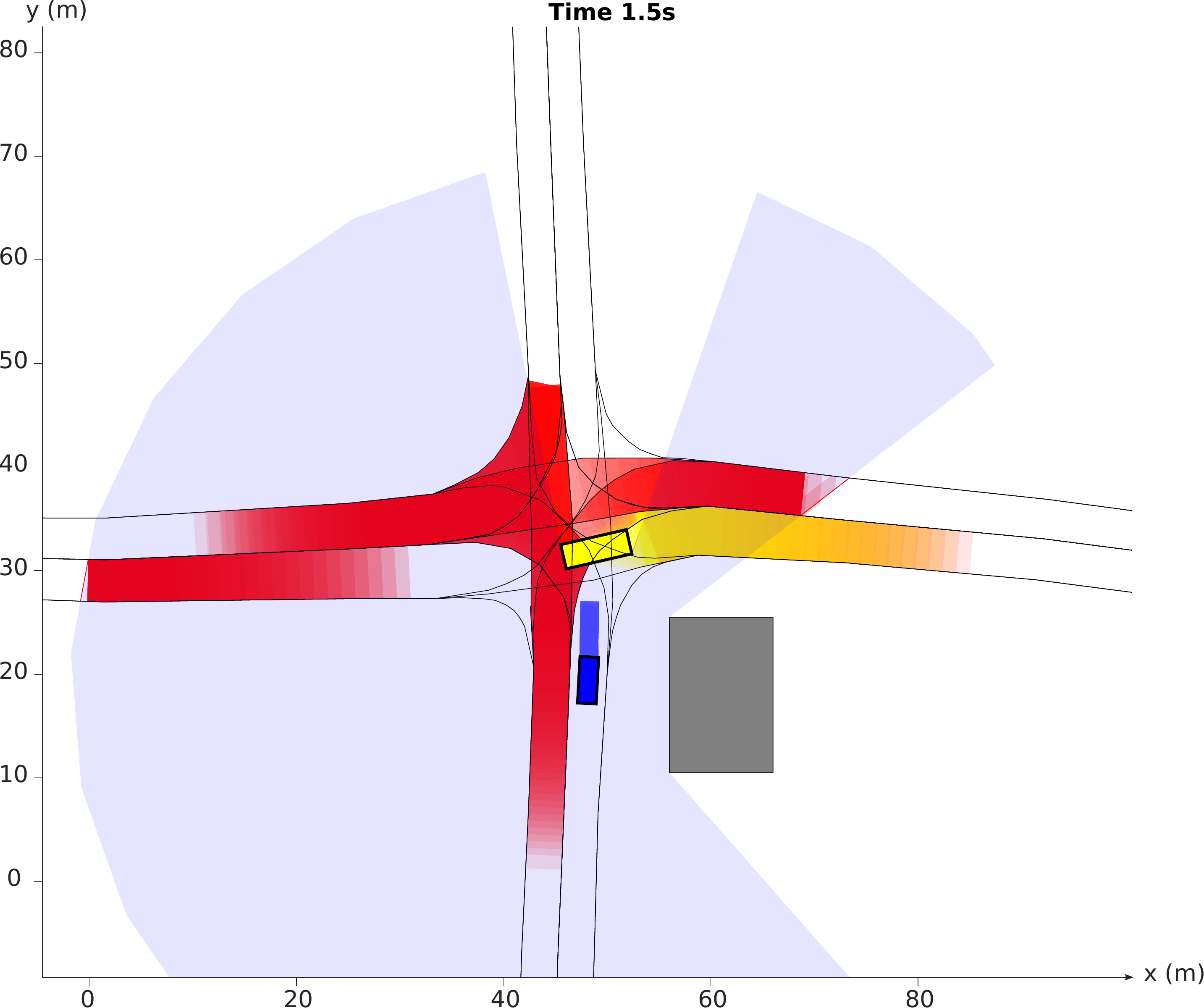}
  \end{subfigure}
  \vspace{.1em}
  \begin{subfigure}[!t]{0.32\textwidth}
    \includegraphics[width=\columnwidth]{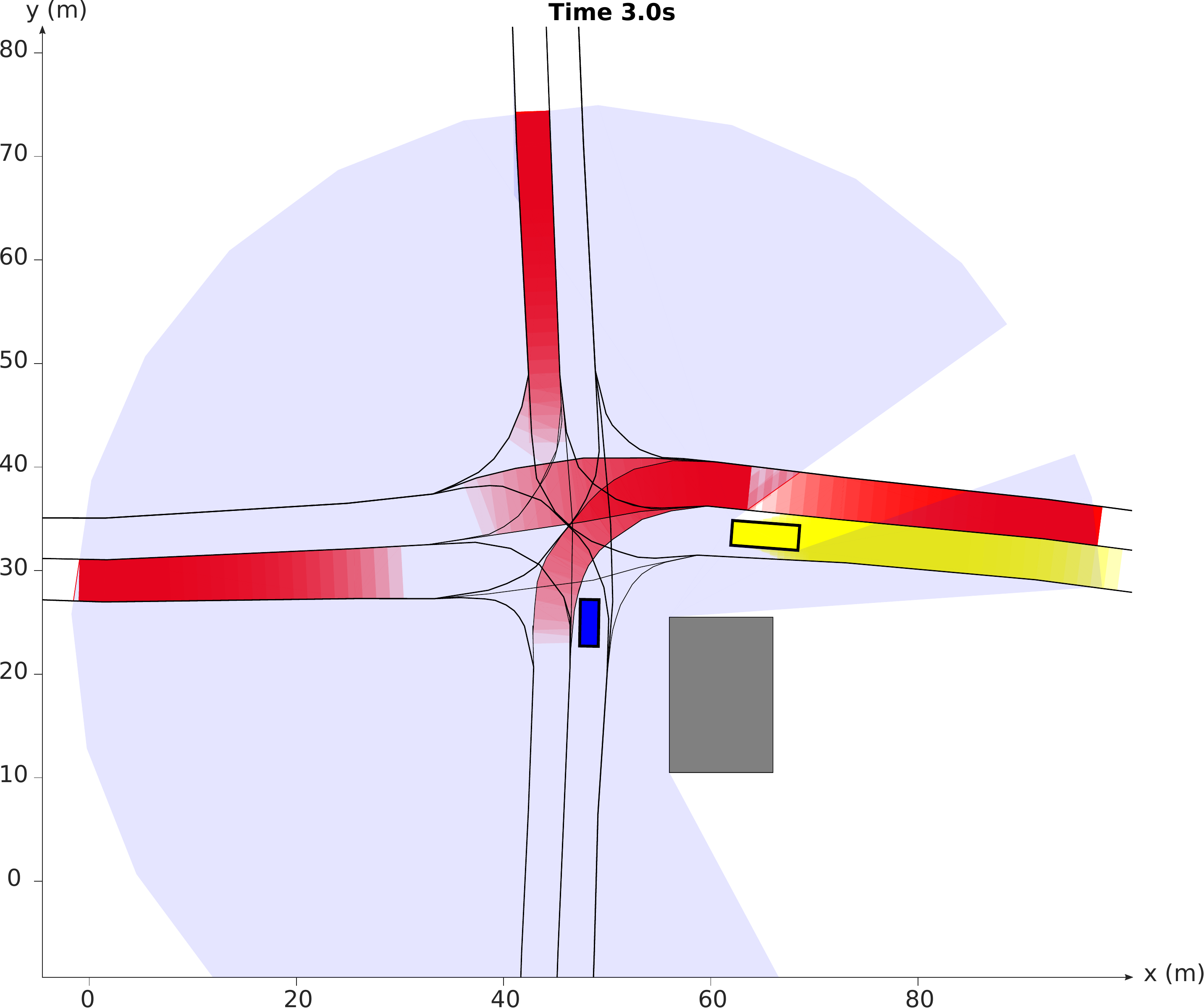}
  \end{subfigure}
  \vspace{.1em}
  \begin{subfigure}[!t]{0.33\textwidth}
    \includegraphics[width=\columnwidth]{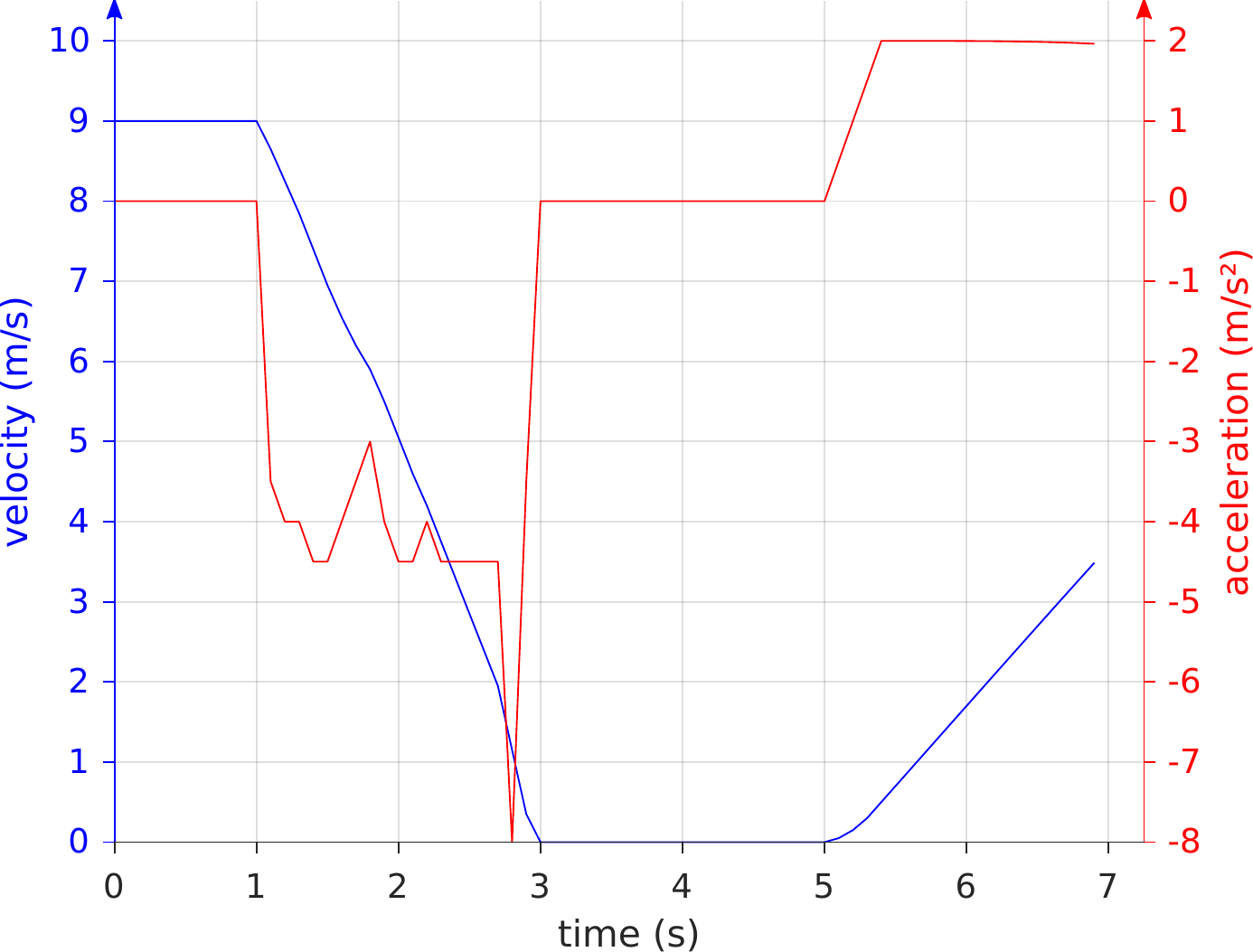}
  \end{subfigure}
  \caption{Problem statement in the first row. Evaluation results on three scenarios below.
  As with many state of the art approaches, the first row shows that a behavior without occlusion-awareness results in a collision.
  With our occupancy prediction of potentially occluded obstacles (red), the same scenario can be passed safely in the second row.
  The third and fourth row show how intersections with occlusions from static and dynamic obstacles can be passed safely.
  }\label{fig:evaluation}
\end{figure*}

\subsection{Merging on a T junction, static occlusion}
\label{subsec:results_T_junction}

The first scenario \textit{DEU\_Ffb\nobreakdash2\_3\_T\nobreakdash1} is a T-junction with a major road leading from east to west with a speed limit of $v_{\text{lim}}=\SI{14}{\mps}$
and a minor road from the north.
One edge of this intersection is occupied by a container, which we doubled in size to dramatize the occlusion effect.

The ego vehicle drives from the north and wants to merge the major road to the west.
The container occludes the eastern arm of the intersection, such that this arm will only be visible well enough around $\SI{2}{\meter}$ before merging into the lane.

The first row in \cref{fig:evaluation} shows the behavior without incorporating occupancies, only based on occupancy predictions of visible obstacles.
As to expect, the ego vehicle does not reduce its velocity in the first $\SI{2.1}{\second}$,
then detects the obstacle and has to decelerate at maximum rate, as it does not find a safe trajectory anymore.
But the emergency braking comes too late and a collision is unavoidable if the other vehicle does not react or
lacks enough reaction time, e.g.\ with a speeding of $\SI{10}{\percent}$ $(\SI{15.4}{\mps})$ or with bad friction, e.g.\ because of a wet road.

The effectiveness of an occlusion aware occupancy prediction with our provided method is shown in the second row of \cref{fig:evaluation}.
The ego vehicle reduces its velocity as soon as the potential trajectory cannot be verified as safe, because it intersects the occupancy of a potentially occluded obstacle.
As in fact there is an obstacle appearing behind the occlusion, the ego vehicle safely comes to a full stop to give way.
Without an obstacle it could safely continue merging after it has decelerated to around $\SI{2.4}{\mps}$ at the point to see far enough to the east.

\subsection{Crossing an X junction, static occlusion}
\label{subsec:results_X_junction_cross}

The second scenario, shown in row 3 of \cref{fig:evaluation} is an intersection in the residential area of Fürstenfeldbruck with a speeding limit of $\SI{40}{\kmh}$.
We model it as an uncontrolled intersection, meaning without traffic lights or stop lines and that the priority-to-the-right rule applies, with $v_{\text{lim}}=\SI{11}{\mps}$.

The ego vehicle comes from the south and wants to cross the intersection.
Similarly to the previous scenario, a static obstacle, in this case a residential building, occludes the easterly lanes.

The simulation without dynamic obstacles shows that the ego vehicle has to slow down significantly, in order to guarantee safety, but can finally pass the intersection.
This is truly an appropriate behavior for such a dangerous setting as such intersections usually feature at least stop lines to force drivers to slow down and have a second look.
The real intersection in Fürstenfeldbruck actually is even regulated by traffic lights.

\subsection{Turning at an X junction, dynamic occlusion}
\label{subsec:results_X_junction_turn}

In the last scenario we use the road geometry from the second Fürstenfeldbruck intersection, but additionally incorporate a dynamic obstacle that passes from west to east.

The ego vehicle wants to cross the intersection to the west, while the dynamic obstacle occludes its view of the easterly lane.
Again the question arises at what time and position it is safe to cross the intersection.

As can be seen in the last row of \cref{fig:evaluation}, the ego vehicle slows down more than without the dynamic obstacle due to the additional occlusion.
Specifically this is the result of assessing every critical sensing field edge with the same prior, i.e.\ the same initial state intervals in velocity and orientation.
It is apparent from the simulation sequence that the east lane is already partly visible before the dynamic obstacle occludes that area.
As a consequence one could derive a better prior for this area based on those observations and continue driving earlier.
We will briefly discuss this possible type of performance improvement in \cref{sec:conclusion}.

Another observation in all scenarios is that the fail-safe trajectory has been activated even without dynamic obstacles.
However, a sophisticated planning method should slow down the vehicle earlier in order to optimize its approaching time and velocity such that the switch to a fail-safe maneuver will rarely be necessary.

Despite those inefficiencies the evaluation shows the need for occlusion-awareness in safety verification and thus highlights the value of our occlusion-aware occupancy prediction.
\section{Conclusions and Future Work}
\label{sec:conclusion}

We motivate this contribution with the purpose of ensuring safety before comfort or traffic efficiency,
especially when facing incomplete environment knowledge.
We therefore characterize the risk from unperceived space as potentially hidden obstacles with an initial state
that is unknown, but can be over-approximated with intervals.
We enhance the reachable set approach introduced in~\cite{althoff_set-based_2016} to
predict occupancy over-approximations of such obstacles.
To show the usefulness and potential of our approach, we implement a proof-of-concept planner
which uses the occupancy prediction to plan provably safe trajectories.
The performance is shown in three intersection scenarios with occlusions from static and dynamic obstacles.
All collisions can be prevented while still moving through traffic fast enough.

Having a method to guarantee safety under occlusions w.r.t.\ an arbitrary safe-state,
further work to improve comfort and efficiency can be done.

Comfort and traffic flow will increase with better prediction, because scenarios with low conflict probability can be approached and passed quicker.
The fail-safe maneuver will still be guaranteed, but only a less comfortable one.
On the other hand the vehicle could decelerate earlier when approaching intersections with high conflict probability.

Similarly, also clever fail-safe velocity profiles can lead to a good compromise between comfortable, but not too conservative behavior.

Tracking of occluded areas would allow to reduce the initial state intervals of hidden obstacles by reasoning.
Hence, the predicted occupancies will be significantly reduced in dynamic scenarios, while still guaranteeing safety.

Finally, we will incorporate the proposed method in our prototype vehicle in the coming months
and analyze the performance under real world conditions.
\section*{Acknowledgements}

The authors thank Daimler AG for the fruitful collaboration and the support for this work.
 
\printbibliography

\end{document}